\def\year{2015}
\documentclass[letterpaper]{article}
\pdfoutput=1
\usepackage{aaai}
\usepackage{times}
\usepackage{helvet}
\usepackage{courier}
\usepackage{algorithm,algorithmic}
\usepackage{multirow}
\usepackage{graphicx}
\usepackage{subfigure}
\usepackage{url}
\begin{document}
%
\title{Robust Image Sentiment Analysis Using Progressively Trained and Domain Transferred Deep Networks}
\author{AAAI Press\\
Association for the Advancement of Artificial Intelligence\\
2275 East Bayshore Road, Suite 160\\
Palo Alto, California 94303\\
}
\author{Quanzeng You \and Jiebo Luo \\
Department of Computer Science \\
University of Rochester \\
Rochester, NY 14623 \\
\{qyou, jluo\}@cs.rochester.edu \\
\And Hailin Jin \and Jianchao Yang \\
Adobe Research \\
345 Park Avenue\\
San Jose, CA 95110\\
\{hljin, jiayang\}@adobe.com\\
}
\maketitle
\begin{abstract}
\begin{quote}
Sentiment analysis of online user generated content is important for many social media analytics tasks. Researchers have largely relied on textual sentiment analysis to develop systems to predict political elections, measure economic indicators, and so on. Recently, social media users are increasingly using images and videos to express their opinions and share their experiences. Sentiment analysis of such large scale visual content can help better extract user sentiments toward events or topics, such as those in image tweets, so that prediction of sentiment from visual content is complementary to textual sentiment analysis. Motivated by the needs in leveraging large scale yet noisy training data to solve the extremely challenging problem of image sentiment analysis, we employ Convolutional Neural Networks (CNN). We first design a suitable CNN architecture for image sentiment analysis. We obtain half a million training samples by using a baseline sentiment algorithm to label Flickr images.  To make use of such noisy machine labeled data, we employ a progressive strategy to fine-tune the deep network. Furthermore, we improve the performance on Twitter images by inducing domain transfer with a small number of manually labeled Twitter images. We have conducted extensive experiments on manually labeled Twitter images. The results show that the proposed CNN can achieve better performance in image sentiment analysis than competing algorithms.
\end{quote}
\end{abstract}

\renewcommand{\algorithmicrequire}{\textbf{Input:}}
\renewcommand{\algorithmicensure}{\textbf{Output:}}
\section{Introduction}
Online social networks are providing more and more convenient services to their users. Today, social networks have grown to be one of the most important sources for people to acquire information on all aspects of their lives. Meanwhile, every online social network user is a contributor to such large amounts of information. Online users love to share their experiences and to express their opinions on virtually all events and subjects.

\begin{figure}[!t]
\begin{centering}
\includegraphics[width=.4\textwidth]{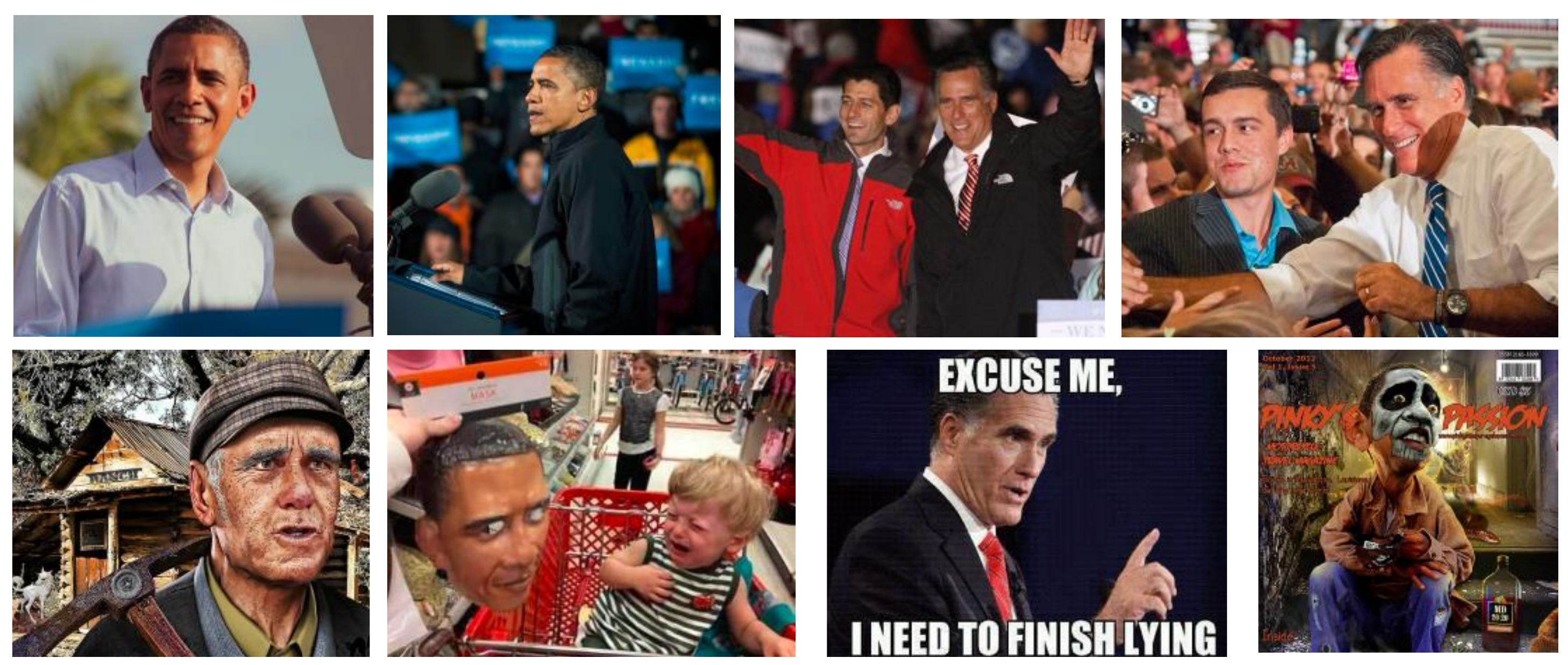}
\caption{Examples of Flickr images related to the 2012 United States presidential election.}
\label{fig:election:imgs}
\end{centering}
\end{figure}

Among the large amount of online user generated data, we are particularly interested in people's opinions or sentiments towards specific topics and events. There have been many works on using online users' sentiments to predict box-office revenues for movies~\cite{asur2010predicting}, political elections~\cite{o2010tweets,tumasjan2010predicting} and economic indicators~\cite{bollen2011twitter,zhang2011predicting}. These works have suggested that online users' opinions or sentiments are closely correlated with our real-world activities. All of these results hinge on accurate estimation of people's sentiments according to their online generated content. Currently all of these works only rely on sentiment analysis from textual content. However, multimedia content, including images and videos, has become prevalent over all online social networks. Indeed, online social network providers are competing with each other by providing easier access to their increasingly powerful and diverse services. \figurename~\ref{fig:election:imgs} shows example images related to the 2012 United States presidential election. Clearly, images in the top and bottom rows convey opposite sentiments towards the two candidates.

\emph{A picture is worth a thousand words}. People with different backgrounds can easily understand the main content of an image or video. Apart from the large amount of easily available visual content, today's computational infrastructure is also much cheaper and more powerful to make the analysis of computationally intensive visual content analysis feasible. In this era of big data, it has been shown that the integration of visual content can provide us more reliable or complementary online social signals~\cite{jin2010wisdom,yuan2013sentribute}.


To the best of our knowledge, little attention has been paid to the sentiment analysis of visual content. Only a few recent works attempted to predict visual sentiment using features from images~\cite{siersdorfer2010analyzing,borth2013large,borth2013sentibank,yuan2013sentribute} and videos~\cite{Morency:2011:TMS:2070481.2070509}. Visual sentiment analysis is extremely challenging. First, image sentiment analysis is inherently more challenging than object recognition as the latter is usually well defined.  Image sentiment involves a much higher level of abstraction and subjectivity in the human recognition process \cite{joshi2011aesthetics}, on top of a wide variety of visual recognition tasks including object, scene, action and event recognition. In order to use supervised learning, it is imperative to collect a large and diverse labeled training set perhaps on the order of millions of images. This is an almost insurmountable hurdle due to the tremendous labor required for image labeling. Second, the learning schemes need to have high generalizability to cover more different domains. However, the existing works use either pixel-level features or a limited number of predefined attribute features, which is difficult to adapt the trained models to images from a different domain.


The deep learning framework enables robust and accurate feature learning, which in turn produces the state-of-the-art performance on digit recognition~\cite{lecun1989backpropagation,hinton2006fast}, image classification~\cite{cirecsan2011flexible,krizhevsky2012imagenet}, musical signal processing~\cite{hamel2010learning} and natural language processing~\cite{maas2011learning}. Both the academia and industry have invested a huge amount of effort in building powerful neural networks. These works suggested that deep learning is very effective in learning robust features in a supervised or unsupervised fashion. Even though deep neural networks may be trapped in local optima~\cite{hinton2010practical,bengio2012practical}, using different optimization techniques, one can achieve the state-of-the-art performance on many challenging tasks mentioned above.

Inspired by the recent successes of deep learning, we are interested in solving the challenging visual sentiment analysis task using deep learning algorithms. For images related tasks, Convolutional Neural Network (CNN) are widely used due to the usage of convolutional layers. It takes into consideration the locations and neighbors of image pixels, which are important to capture useful features for visual tasks. Convolutional Neural Networks~\cite{lecun1998gradient,cirecsan2011flexible,krizhevsky2012imagenet} have been proved very powerful in solving computer vision related tasks. We intend to find out whether applying CNN to visual sentiment analysis provides advantages over using a predefined collection of low-level visual features or visual attributes, which have been done in prior works.

To that end, we address in this work two major challenges: 1) how to learn with large scale weakly labeled training data, and 2) how to generalize and extend the learned model across domains. In particular, we make the following contributions.
\begin{itemize}
\item{We develop an effective deep convolutional network architecture for visual sentiment analysis. Our architecture employs two convolutional layers and several fully connected layers for the prediction of visual sentiment labels.}
\item{Our model attempts to address the weakly labeled nature of the training image data, where such labels are machine generated, by leveraging a progressive training strategy and a domain transfer strategy to fine-tune the neural network. Our evaluation results suggest that this strategy is effective for improving the performance of neural network in terms of generalizability.}
\item{In order to evaluate our model as well as competing algorithms, we build a large manually labeled visual sentiment dataset using Amazon Mechanical Turk. This dataset will be released to the research community to promote further investigations on visual sentiment.}
\end{itemize}
\section{Related Work}
In this section, we review literature closely related to our study on visual sentiment analysis, particularly in sentiment analysis and Convolutional Neural Networks.
\subsection{Sentiment Analysis}

Sentiment analysis is a very challenging task~\cite{liu2003building,li2010micro}. Researchers from natural language processing and information retrieval have developed different approaches to solve this problem, achieving promising or satisfying results~\cite{pang2008opinion}. In the context of social media, there are several additional unique challenges. First, there are huge amounts of data available. Second, messages on social networks are by nature informal and short. Third, people use not only textual messages, but also images and videos to express themselves.

Tumasjan et al.~\shortcite{tumasjan2010predicting} and Bollen et al.~\shortcite{bollen2011modeling} employed pre-defined dictionaries for measuring the sentiment level of Tweets. The volume or percentage of sentiment-bearing words can produce an estimate of the sentiment of one particular tweet. Davidov et al.~\shortcite{davidov2010enhanced} used the weak labels from a large amount of Tweets. In contrast, they manually selected hashtags with strong positive and negative sentiments and ASCII smileys are also utilized to label the sentiments of tweets. Furthermore, Hu et al.~\shortcite{hu2013unsupervised} incorporated social signals into their unsupervised sentiment analysis framework. They defined and integrated both emotion indication and correlation into a framework to learn parameters for their sentiment classifier.
\begin{figure*}[!ht]
\begin{minipage}{0.9\linewidth}
\begin{centering}
\includegraphics[width=.7\textwidth]{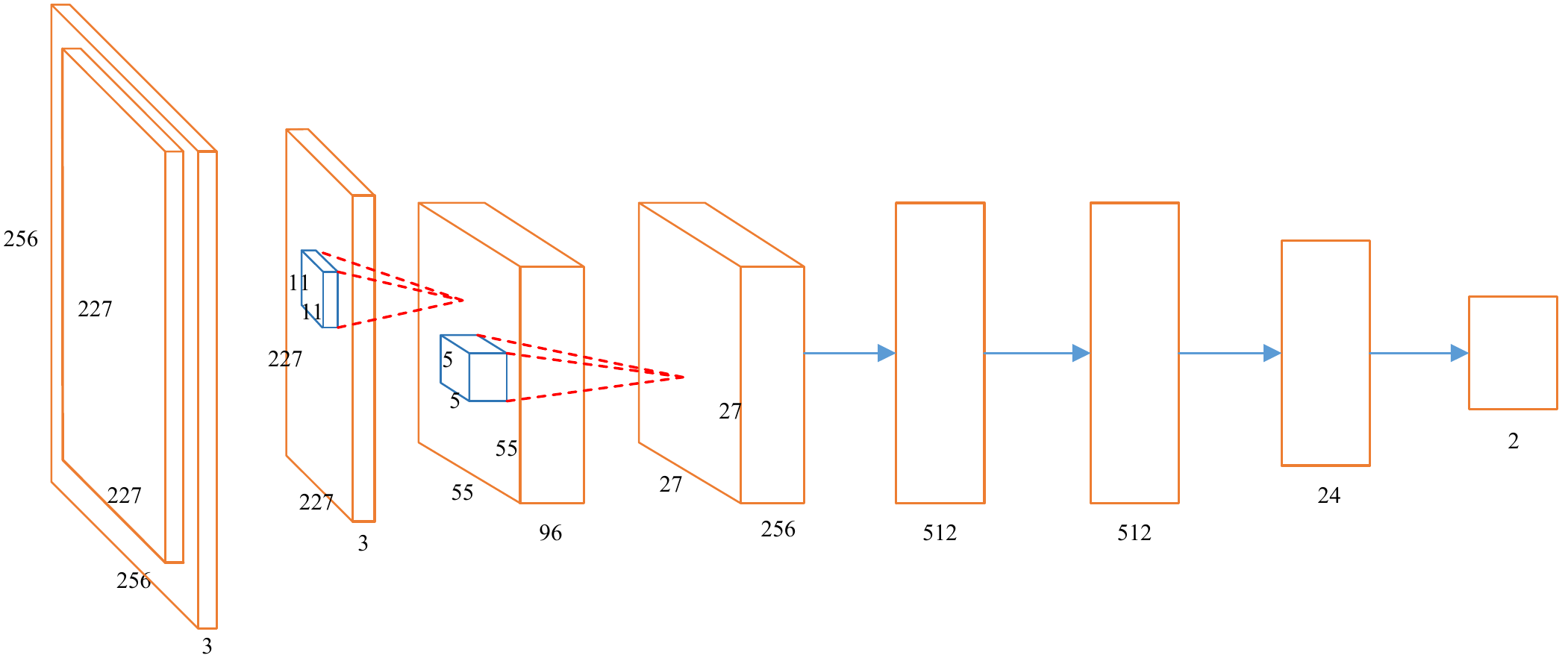}
\caption{Convolutional Neural Network for Visual Sentiment Analysis.}
\label{fig:dl:sa}
\end{centering}
\end{minipage}
\end{figure*}

There are also several recent works on visual sentiment analysis. Siersdorfer et al.~\shortcite{siersdorfer2010analyzing} proposes a machine learning algorithm to predict the sentiment of images using pixel-level features. Motivated by the fact that sentiment involves high-level abstraction, which may be easier to explain by objects or attributes in images, both~\cite{borth2013sentibank} and~\cite{yuan2013sentribute} propose to employ visual entities or attributes as features for visual sentiment analysis. In~\cite{borth2013sentibank}, 1200 adjective noun pairs (ANP), which may correspond to different levels of different emotions, are extracted. These ANPs are used as queries to crawl images from Flickr. Next, pixel-level features of images in each ANP are employed to train 1200 ANP detectors. The responses of these 1200 classifiers can then be considered as mid-level features for visual sentiment analysis. The work in~\cite{yuan2013sentribute} employed a similar mechanism. The main difference is that 102 scene attributes are used instead.
\subsection{Convolutional Neural Networks}
Convolutional Neural Networks (CNN) have been very successful in document recognition~\cite{lecun1998gradient}. CNN typically consists of several convolutional layers and several fully connected layers. Between the convolutional layers, there may also be pooling layers and normalization  layers. CNN is a supervised learning algorithm, where parameters of different layers are learned through back-propagation. Due to the computational complexity of CNN, it has only be applied to relatively small images in the literature. Recently, thanks to the increasing computational power of GPU, it is now possible to train a deep convolutional neural network on a large scale image dataset~\cite{krizhevsky2012imagenet}. Indeed, in the past several years, CNN has been successfully applied to scene parsing~\cite{grangier2009deep}, feature learning~\cite{lecun2010convolutional}, visual recognition~\cite{kavukcuoglu2010learning} and image classification~\cite{krizhevsky2012imagenet}. In our work, we intend to use CNN to learn features which are useful for visual sentiment analysis.
\section{Visual Sentiment Analysis}
We propose to develop a suitable convolutional neural network architecture for visual sentiment analysis. Moreover, we employ a progressive training strategy that leverages the training results of convolutional neural network to further filter out (noisy) training data. The details of the proposed framework will be described in the following sections.

\subsection{Visual Sentiment Analysis with regular CNN}
CNN has been proven to be effective in image classification tasks, e.g., achieving the state-of-the-art performance in ImageNet Challenge~\cite{krizhevsky2012imagenet}. Visual sentiment analysis can also be treated as an image classification problem. It may seem to be a much easier problem than image classification from ImageNet (2 classes vs. 1000 classes in ImageNet). However, visual sentiment analysis is quite challenging because sentiments or opinions correspond to high level abstractions from a given image. This type of high level abstraction may require viewer's knowledge beyond the image content itself. Meanwhile, images in the same class of ImageNet mainly contain the same type of object. In sentiment analysis, each class contains much more diverse images. It is therefore extremely challenging to discover features which can distinguish much more diverse classes from each other. In addition, people may have totally different sentiments over the same image. This adds difficulties to not only our classification task, but also the acquisition of labeled images. In other words, it is nontrivial to obtain highly reliable labeled instances, let alone a large number of them. Therefore, we need a supervised learning engine that is able to tolerate a significant level of noise in the training dataset.

The architecture of the CNN we employ for sentiment analysis is shown in~\figurename~\ref{fig:dl:sa}. Each image is resized to $256\times 256$ (if needed, we employ center crop, which first resizes the shorter dimension to 256 and then crops the middle section of the resized image). The resized images are processed by two convolutional layers. Each convolutional layer is also followed by max-pooling layers and normalization layers. The first convolutional layer has $96$ kernels of size $11\times 11 \times 3$ with a stride of $4$ pixels. The second convolutional layer has $256$ kernels of size $5 \times 5$ with a stride of $2$ pixels. Furthermore, we have four fully connected layers. Inspired by~\cite{GulcehreCPB13}, we constrain the second to last fully connected layer to have $24$ neurons. According to the Plutchik's wheel of emotions~\cite{plutchik1984emotions}, there are a total of $24$ emotions belonging to two categories: positive emotions and negative emotions. Intuitively. we hope these 24 nodes may help the network to learn the $24$ emotions from a given image and then classify each image into positive or negative class according to the responses of these $24$ emotions.

The last layer is designed to learn the parameter $w$ by maximizing the following conditional log likelihood function ($x_i$ and $y_i$ are the feature vector and label for the $i$-th instance respectively):
\begin{equation}
l(w) = \sum_{i=1}^n \ln p(y_i = 1|x_i, w) + ( 1 - y_i)\ln p(y_i = 0 | x_i, w)
\label{eqn:loglikely}
\end{equation}
where
\begin{equation}
p(y_i|x_i,w) = \frac{\exp(w_0 +\sum_{j=1}^k w_j x_{ij})^{y_i}}{1 + \exp(w_0 +\sum_{j=1}^k w_j x_{ij})^{y_i}}
\label{eqn:func}
\end{equation}
\begin{figure*}[!t]
\begin{centering}
\includegraphics[width=.8\textwidth]{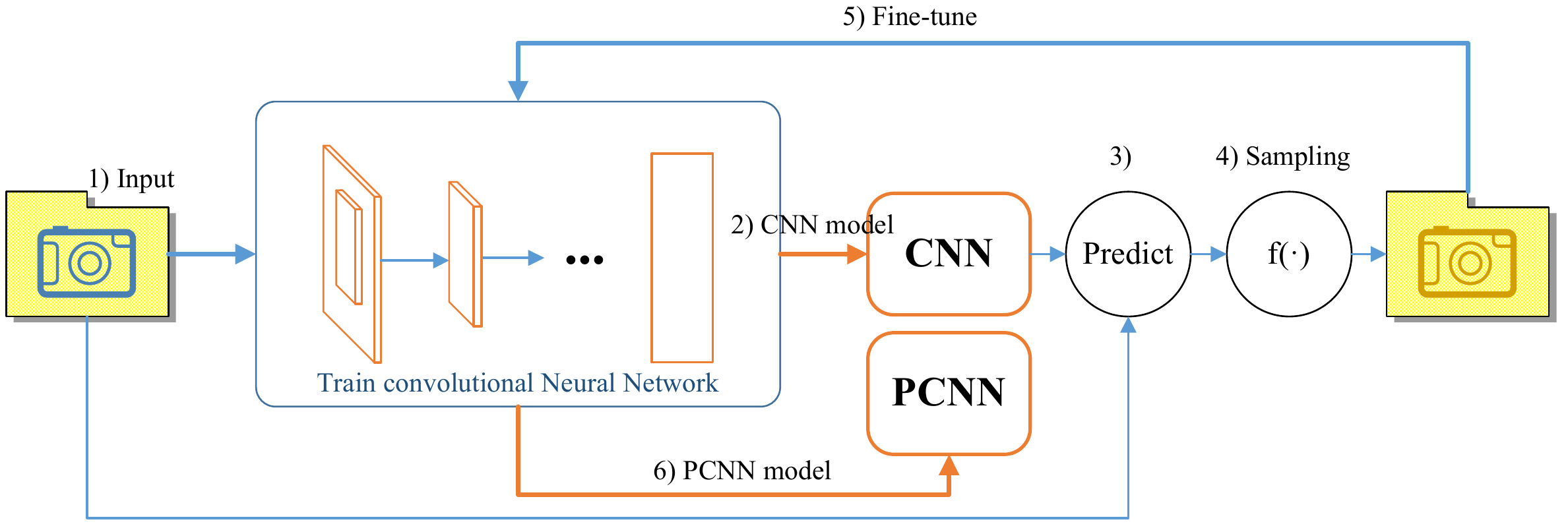}
\caption{Progressive CNN (PCNN) for visual sentiment analysis.}
\label{fig:pcnn}
\end{centering}
\end{figure*}
\subsection{Visual Sentiment Analysis with Progressive CNN}
Since the images are weakly labeled, it is possible that the neural network can get stuck in a bad local optimum. This may lead to poor generalizability of the trained neural network. On the other hand, we found that the neural network is still able to correctly classify a large proportion of the training instances. In other words, the neural network has learned knowledge to distinguish the training instances with relatively distinct sentiment labels. Therefore, we propose to progressively select a subset of the training instances to reduce the impact of noisy training instances. \figurename~\ref{fig:pcnn} shows the overall flow of the proposed progressive CNN (PCNN). We first train a CNN on Flickr images. Next, we select training samples according to the prediction score of the trained model on the training data itself. Instead of training from the beginning, we further fine-tune the trained model using these newly selected, and potentially cleaner training instances. This fine-tuned model will be our final model for visual sentiment analysis.

\begin{algorithm}[h!]
\caption{Progressive CNN training for Visual Sentiment Analysis}
\label{alg:pcnn}
\begin{algorithmic}[1]
\REQUIRE $X=\{x_1,x_2,\dots,x_n\}$ a set of images of size $256\times 256$ \\
\quad \ \ $Y=\{y_1, y_2, \dots, y_n\}$ sentiment labels of $X$
\STATE Train convolutional neural network {\em{CNN}} with input $X$ and $Y$
\STATE Let $S\in R^{n\times 2}$ be the sentiment scores of $X$ predicted using {\em{CNN}}
\FOR { $s_i \in S$}
    \STATE Delete $x_i$ from $X$ with probability $p_i$ (Eqn.(\ref{eqn:sampling}))
\ENDFOR
\STATE Let $X'\subset X$ be the remaining training images, $Y'$ be their sentiment labels
\STATE Fine-tune {\em{CNN}} with input $X'$ and $Y'$ to get {\em{PCNN}}
\RETURN {\em{PCNN}}
\end{algorithmic}
\end{algorithm}

In particular, we employ a probabilistic sampling algorithm to select the new training subset. The intuition is that we want to keep instances with distinct sentiment scores between the two classes with a high probability, and conversely remove instances with similar sentiment scores for both classes with a high probability. Let $s_i = (s_{i1}, s_{i2})$ be the prediction sentiment scores for the two classes of instance $i$. We choose to remove the training instance $i$ with probability $p_i$ given by Eqn.(\ref{eqn:sampling}). Algorithm~\ref{alg:pcnn} summarizes the steps of the proposed framework.
\begin{equation}
p_i =  \max\left(0, 2 - \exp(|s_{i1} - s_{i2}|)\right)
\label{eqn:sampling}
\end{equation}
When the difference between the predicted sentiment scores of one training instance are large enough, this training instance will be kept in the training set. Otherwise, the smaller the difference between the predicted sentiment scores become, the larger the probability of this instance being removed from the training set.

\section{Experiments}
We choose to use the same half million Flickr images from SentiBank\footnote{http://visual-sentiment-ontology.appspot.com/} to train our Convolutional Neural Network. These images are only weakly labeled since each image belongs to one adjective noun pair (ANP). There are a total of 1200 ANPs. According to the Plutchik's Wheel of Emotions~\cite{plutchik1984emotions}, each ANP is generated by the combination of adjectives with strong sentiment values and nouns from tags of images and videos~\cite{borth2013large}. These ANPs are then used as queries to collect related images for each ANP. The released SentiBank contains 1200 ANPs with about half million Flickr images. We train our convolutional neural network mainly on this image dataset.
We implement the proposed architecture of CNN on the publicly available implementation Caffe~\cite{Jia13caffe}.
All of our experiments are evaluated on a Linux X86\_64 machine with 32G RAM and two NVIDIA GTX Titan GPUs.

\subsection{Comparisons of different CNN architectures}
The architecture of our model is shown in~\figurename~\ref{fig:dl:sa}. However, we also evaluate other architectures for the visual sentiment analysis task. \tablename~\ref{tab:cnns} summarizes the performance of different architectures on a randomly chosen Flickr testing dataset. In~\tablename~\ref{tab:cnns}, $i$CONV-$j$FC indicates that there are $i$ convolutional layers and $j$ fully connected layers in the architecture. The model in~\figurename~\ref{fig:dl:sa} shows slightly better performance than other models in terms of F1 and accuracy. In the following experiments, we mainly focus on the evaluation of CNN using the architecture in~\figurename~\ref{fig:dl:sa}.
\begin{table}[!tbh]
    \caption{Summary of performance of different architectures on randomly chosen testing data.}
    \label{tab:cnns}
    \centering
    \small
    \begin{tabular}{|l|c|c|c|c|}
        \hline
        Architecture & Precision & Recall & F1 & Accuracy\\
        \hline
        3CONV-4FC & 0.679 & 0.845 & 0.753 & 0.644 \\
        \hline
        3CONV-2FC & 0.69 & 0.847 & 0.76 & 0.657 \\
        \hline
        2CONV-3FC & 0.679 & 0.874 & 0.765 & 0.654 \\
        \hline
        2CONV-4FC & 0.688 & 0.875 & 0.77 & 0.665 \\
        \hline
    \end{tabular}
\end{table}

\subsection{Baselines}
We compare the performance of PCNN with three other baselines or competing algorithms for image sentiment classification.
\subsubsection{Low-level Feature-based}
Siersdorfer et al.~\shortcite{siersdorfer2010analyzing} defined both global and local visual features. Specifically, the global color histograms (\textbf{GCH}) features consist of 64-bin RGB histogram. The local color histogram features (\textbf{LCH}) first divided the image into 16 blocks and used the 64-bin RGB histogram for each block. They also employed SIFT features to learn a visual word dictionary. Next, they defined bag of visual word features (\textbf{BoW}) for each image.
\subsubsection{Mid-level Feature-based}
Damian et al.~\shortcite{borth2013sentibank,borth2013large} proposed a framework to build visual sentiment ontology and~\textbf{SentiBank} according to the previously discussed 1200 ANPs. With the trained 1200 ANP detectors, they are able to generate 1200 responses for any given test image using these pre-trained 1200 ANP detectors. A sentiment classifier is built on top of these mid-level features according to the sentiment label of training images. \textbf{Sentribute}~\cite{yuan2013sentribute} also employed mid-level features for sentiment prediction. However, instead of using adjective noun pairs, they employed scene-based attributes~\cite{Patterson2012SunAttributes} to define the mid-level features.

\subsection{Deep Learning on Flickr Dataset}
\begin{figure}[!ht]
\begin{centering}
\subfigure[Filters learned from CNN]{
            \includegraphics[width=.43\textwidth]{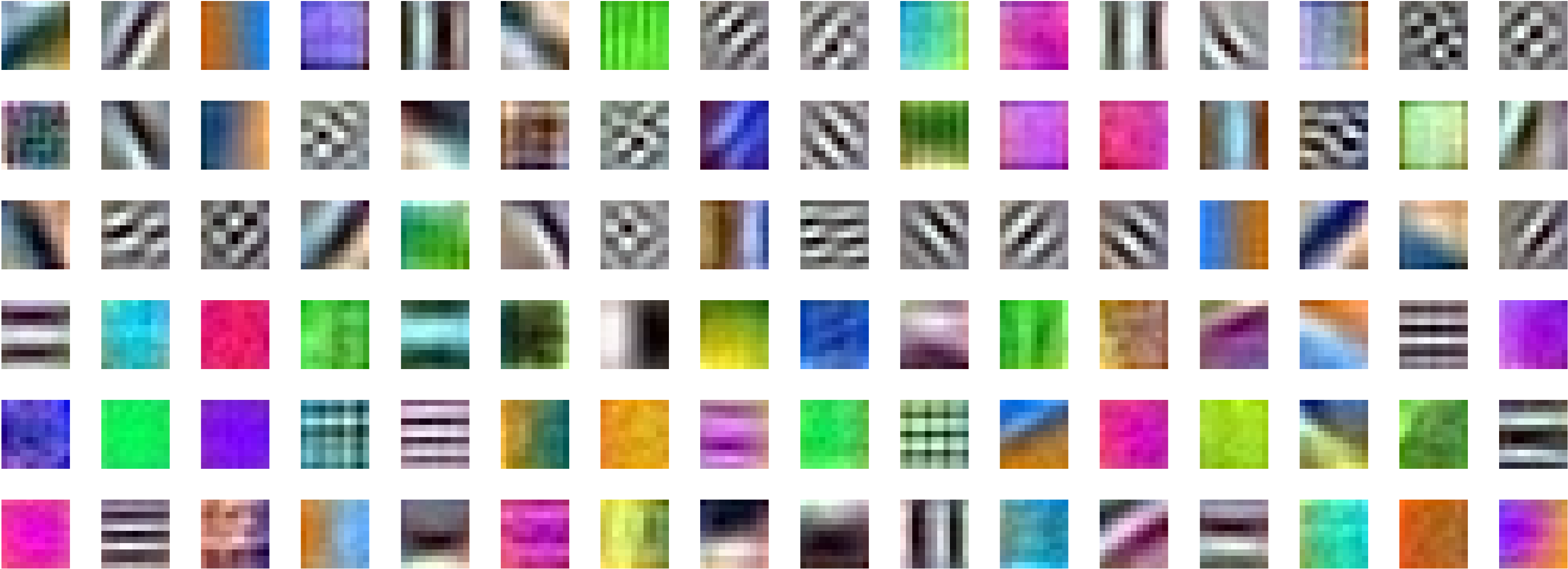}
         \label{fig:filtering:cnn}
}
\subfigure[Filters learned from PCNN]{
            \includegraphics[width=.43\textwidth]{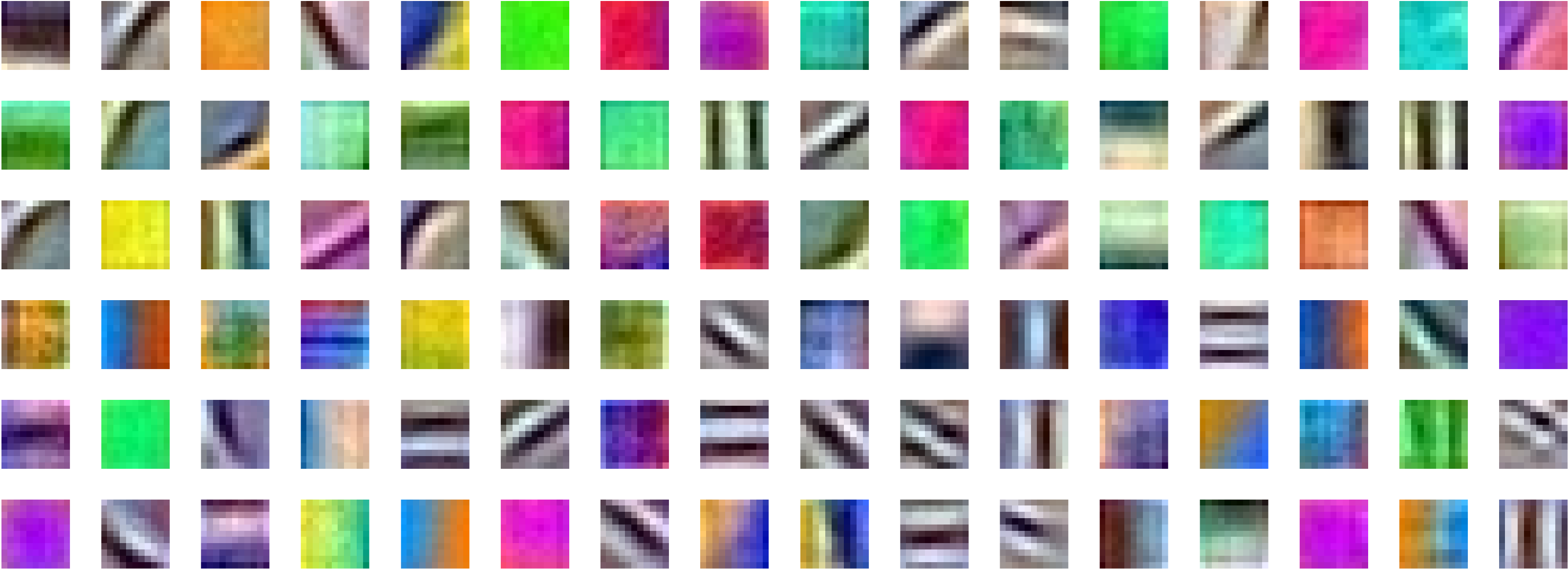}
         \label{fig:filtering:pcnn}
}
\caption{Filters of the first convolutional layer. }
\label{fig:filtering}
\end{centering}
\end{figure}

We randomly choose $90\%$ images from the half million Flickr images as our training dataset. The remaining $10\%$ images are our testing dataset. We train the convolutional neural network with 300,000 iterations of mini-batches (each mini-batch contains 256 images). We employ the sampling probability in Eqn.(\ref{eqn:sampling}) to filter the training images according to the prediction score of CNN on its training data. In the fine-tuning stage of PCNN, we run another 100,000 iterations of mini-batches using the filtered training dataset. \tablename~\ref{tab:flickr:stat} gives a summary of the number of data instances in our experiments.
\begin{table}[!htb]
    \caption{Statistics of the number of Flickr image dataset.}
    \label{tab:flickr:stat}
    \centering
    \begin{tabular}{|l|c|c|c|}
        \hline
        Models & training & testing & \# of iterations\\
        \hline
        CNN & 401,739 & 44,637 & 300,000\\
        \hline
        PCNN& 369,828  & 44,637 & 100,000\\
        \hline
    \end{tabular}
\end{table}
\begin{table}[!hbtp]
\centering{
\caption{Performance on the Testing Dataset by CNN and PCNN.}
\label{tab:flickr:testing}
\small
\begin{tabular}{*{9}{|l}|} \hline
Algorithm & Precision & Recall & F1 & Accuracy \\ \hline
CNN & 0.714 & 0.729 & 0.722 &  0.718 \\ \hline
PCNN &  \textbf{0.759} & \textbf{0.826} & \textbf{0.791} & \textbf{0.781}\\ \hline
\end{tabular}
}
\end{table}
\figurename~\ref{fig:filtering} shows the filters learned in the first convolutional layer of CNN and PCNN, respectively. There are some differences between~\ref{fig:filtering:cnn} and ~\ref{fig:filtering:pcnn}. While it is somewhat inconclusive that the neural networks have reached a better local optimum, at least we can conclude that the fine-tuning stage using a progressively cleaner training dataset has prompted the neural networks to learn different knowledge. Indeed, the evaluation results suggest that this fine-tuning leads to the improvement of performance.

\tablename~\ref{tab:flickr:testing} shows the performance of both CNN and PCNN on the $10\%$ randomly chosen testing data. PCNN outperformed CNN in terms of \emph{Precision}, \emph{Recall}, \emph{F1} and \emph{Accuracy}. The results in \tablename~\ref{tab:flickr:testing} and the filters from \figurename~\ref{fig:filtering} shows that the fine-tuning stage of PCNN can help the neural network to search for a better local optimum.
\subsection{Twitter Testing Dataset}

We also built a new image dataset from image tweets. Image tweets refer to those tweets that contain images. We built a total of 1269 images as our candidate testing images. We employed crowd intelligence, Amazon Mechanical Turk (AMT), to generate sentiment labels for these testing images, in a similar fashion to \cite{borth2013large}. We recruited 5 AMT workers for each of the candidate image. \tablename~\ref{tab:amt:twitter} shows the statistics of the labeling results from the Amazon Mechanical Turk. In the table, ``five agree" indicates that all the 5 AMT workers gave the same sentiment label for a given image. Only a small portion of the images, 153 out of 1269, had significant disagreements between the 5 workers (3 vs. 2).
\begin{table}[!tbh]
    \caption{Summary of AMT labeled results for the Twitter testing dataset.}
    \label{tab:amt:twitter}
    \centering
    \small
    \begin{tabular}{|l|c|c|c|}
    \hline
\multirow{2}{*}{Sentiment } & \multirow{2}{*}{Five Agree}& \multicolumn{1}{|c|}{\multirow{2}{.1\textwidth}{At Least Four Agree }}&\multicolumn{1}{|c|}{\multirow{2}{.1\textwidth}{At Least Three Agree }} \\
& & & \\ \hline
        Positive & 581 & 689 & 769\\
        \hline
        Negative & 301 & 427 & 500\\
        \hline
        Sum & 882 & 1116 & 1269 \\
        \hline
    \end{tabular}
\end{table}
\begin{table*}[!hbtp]
\centering{
\caption{Performance of different algorithms on the Twitter image dataset (Acc stands for Accuracy).}
\label{tab:cnn:twitter}
\small
\begin{tabular}{*{13}{|l}|} \hline
\multicolumn{1}{|c|}{\multirow{2}{*}{ Algorithms }} & \multicolumn{4}{|c|}{Five Agree} &
\multicolumn{4}{|c|}{At Least Four Agree} & \multicolumn{4}{|c|}{At Least Three Agree} \\
 \cline{2-13}
& \multicolumn{1}{|c|}{Precision} & \multicolumn{1}{|c|}{Recall} & \multicolumn{1}{|c|}{F1} & \multicolumn{1}{|c|}{Acc}& \multicolumn{1}{|c|}{Precision} & \multicolumn{1}{|c|}{Recall} & \multicolumn{1}{|c|}{F1} & \multicolumn{1}{|c|}{Acc}& \multicolumn{1}{|c|}{Precision} & \multicolumn{1}{|c|}{Recall} & \multicolumn{1}{|c|}{F1} & \multicolumn{1}{|c|}{Acc}\\ \hline
CNN & 0.749 & 0.869         & 0.805             & 0.722 & 0.707 & 0.839 & 0.768 & 0.686 & 0.691 & \textbf{0.814} & 0.747 & 0.667 \\ \hline
PCNN & \textbf{0.77} & \textbf{0.878} & \textbf{0.821} & \textbf{0.747} & \textbf{0.733} & \textbf{0.845} & \textbf{0.785} & \textbf{0.714} &\textbf{0.714}& 0.806 & \textbf{0.757} & \textbf{0.687} \\ \hline
\end{tabular}
}
\end{table*}
We evaluate the performance of Convolutional Neural Networks on this manually labeled image dataset according to the model trained on Flickr images.~\tablename~\ref{tab:cnn:twitter} shows the performance of the two frameworks. Not surprisingly, both models perform better on the less ambiguous image set (``five agree" by AMT). Meanwhile, PCNN shows better performance than CNN on all the three labeling sets in terms of both F1 and accuracy. This suggests that the fine-tuning stage of CNN effectively improves the generalizability extensibility of the neural networks.
\begin{figure}[!ht]
\begin{centering}
    \includegraphics[width=.42\textwidth]{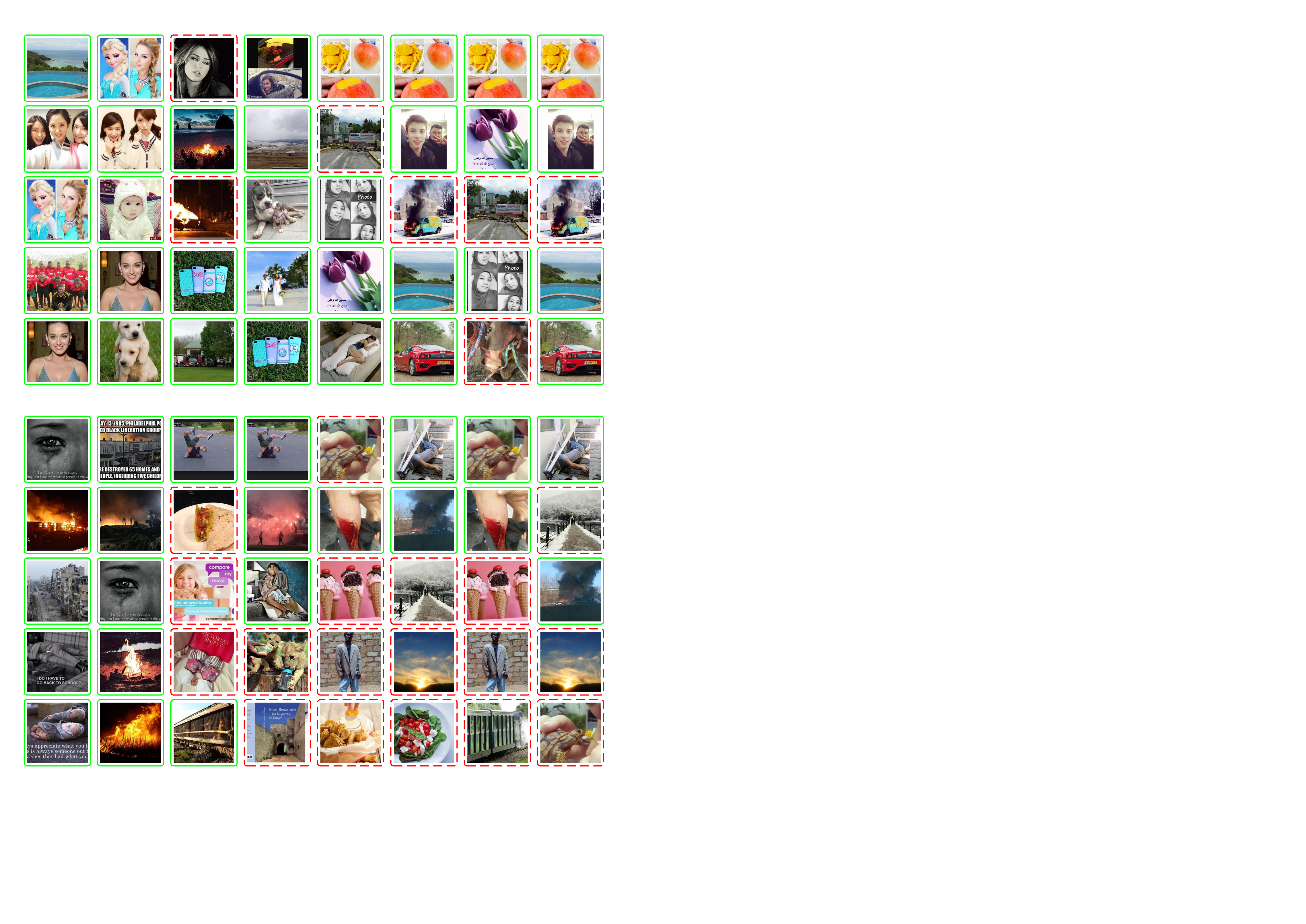}
\caption{Positive (top block) and Negative (bottom block) examples. Each column shows the negative example images for each algorithm (PCNN, CNN, Sentribute, Sentibank, GCH, LCH, GCH+BoW, LCH+BoW). The images are ranked by the prediction score from top to bottom in a decreasing order.}
\end{centering}
\label{fig:neg}
\end{figure}
\subsection{Transfer Learning}
Half million Flickr images are used in our CNN training. The features learned are generic features on these half million images. \tablename~\ref{tab:cnn:twitter} shows that these generic features also have the ability to predict visual sentiment of images from other domains. The question we ask is whether we can further improve the performance of visual sentiment analysis on Twitter images by inducing transfer learning. In this section, we conduct experiments to answer this question.

The users of Flickr are more likely to spend more time on taking high quality pictures. Twitter users are likely to share the moment with the world. Thus, most of the Twitter images are casually taken snapshots. Meanwhile, most of the images are related to current trending topics and personal experiences, making the images on Twitter much diverse in content as well as quality.

In this experiment, we fine-tune the pre-trained neural network model in the following way to achieve transfer learning. We randomly divide the Twitter images into 5 equal partitions. Every time, we use 4 of the 5 partitions to fine-tune our pre-trained model from the half million Flickr images and evaluate the new model on the remaining partition. The averaged evaluation results are reported. The algorithm is detailed in Algorithm~\ref{alg:tl}.
\begin{algorithm}[h!]
\caption{Transfer Learning to fine-tune CNN}
\label{alg:tl}
\begin{algorithmic}[1]
\REQUIRE $X=\{x_1,x_2,\dots,x_n\}$ a set of images of size $256\times 256$ \\
\quad \ \ $Y=\{y_1, y_2, \dots, y_n\}$ sentiment labels of $X$ \\
\quad \ \ Pre-trained CNN model $M$
\STATE Randomly partition $X$ and $Y$ into $5$ equal groups $\{(X_1, Y_1), \dots, (X_5, Y_5)\}$.
\FOR {$i$ from 1 to 5}
    \STATE Let $(X', Y') = (X,Y) - (X_i, Y_i)$
    \STATE Fine-tune $M$ with input $(X',Y')$ to obtain model $M_i$
    \STATE Evaluate the performance of $M_i$ on $(X_i, Y_i)$
\ENDFOR
\RETURN The averaged performance of $M_i$ on $(X_i, Y_i)$ ($i$ from 1 to 5)
\end{algorithmic}
\end{algorithm}

Similar to~\cite{borth2013large}, we also employ 5-fold cross-validation to evaluate the performance of all the baseline algorithms. \tablename~\ref{tab:cmp} summarizes the averaged performance results of different baseline algorithms and our two CNN models. Overall, both CNN models outperform the baseline algorithms. In the baseline algorithms, Sentribute gives slightly better results than the other two baseline algorithms. Interestingly, even the combination of using low-level features local color histogram (LCH) and bag of visual words (BoW) shows better results than SentiBank on our Twitter dataset.
\begin{table*}[!tp]
\centering{
\caption{5-Fold Cross-Validation Performance of different algorithms on the Twitter image dataset. Note that compared with Table~\ref{tab:cnn:twitter}, both fine-tuned CNN models have been improved due to domain transfer learning (Acc stands for Accuracy).}
\label{tab:cmp}
\small
\begin{tabular}{*{13}{|l}|} \hline
\multicolumn{1}{|c|}{\multirow{2}{*}{ Algorithms }} & \multicolumn{4}{|c|}{Five Agree} &
\multicolumn{4}{|c|}{At Least Four Agree} & \multicolumn{4}{|c|}{At Least Three Agree} \\
 \cline{2-13}
& \multicolumn{1}{|c|}{Precision} & \multicolumn{1}{|c|}{Recall} & \multicolumn{1}{|c|}{F1} & \multicolumn{1}{|c|}{Acc}& \multicolumn{1}{|c|}{Precision} & \multicolumn{1}{|c|}{Recall} & \multicolumn{1}{|c|}{F1} & \multicolumn{1}{|c|}{Acc}& \multicolumn{1}{|c|}{Precision} & \multicolumn{1}{|c|}{Recall} & \multicolumn{1}{|c|}{F1} & \multicolumn{1}{|c|}{Acc}\\ \hline
GCH & 0.708 & 0.888 & 0.787 & 0.684 & 0.687 & 0.84 & 0.756 & 0.665 & 0.678 & \textbf{0.836} & 0.749 & 0.66 \\ \hline
LCH & 0.764 & 0.809 & 0.786 & 0.71 &  0.725 & 0.753 & 0.739 & 0.671& 0.716 & 0.737 & 0.726 & 0.664 \\ \hline
GCH + BoW & 0.724 & 0.904 & 0.804 & 0.71 &0.703 & 0.849 & 0.769 & 0.685 & 0.683 & \textbf{0.835} & 0.751 & 0.665 \\ \hline
LCH + BoW & 0.771 & 0.811 & 0.79 & 0.717 & 0.751 & 0.762 & 0.756 & 0.697 & 0.722 & 0.726 & 0.723 & 0.664 \\ \hline
SentiBank & 0.785  & 0.768 & 0.776 & 0.709 &  0.742 &0.727 & 0.734 & 0.675& 0.720 & 0.723 & 0.721 & 0.662 \\ \hline
Sentribute & 0.789 & 0.823  & 0.805 & 0.738  & 0.75 & 0.792 & 0.771 & 0.709 & 0.733 & 0.783 & 0.757 & 0.696 \\ \hline
CNN &  \textbf{0.795} & \textbf{0.905} & \textbf{0.846} & \textbf{0.783} & \textbf{0.773}  &\textbf{0.855} & \textbf{0.811} & \textbf{0.755} & \textbf{0.734} & \textbf{0.832} & \textbf{0.779} & \textbf{0.715} \\ \hline
PCNN & \textbf{0.797} & \textbf{0.881} & \textbf{0.836} & \textbf{0.773} & \textbf{0.786} & \textbf{0.842} & \textbf{0.811} & \textbf{0.759} & \textbf{0.755} & \textbf{0.805} & \textbf{0.778} & \textbf{0.723} \\ \hline
\end{tabular}
}
\end{table*}
Both fine-tuned CNN models have been improved. This improvement is significant given that we only use four fifth of the 1269 images for domain adaptation. Both neural network models have similar performance on all the three sets of the Twitter testing data. This suggests that the fine-tuning stage helps both models to find a better local minimum. In particular, the knowledge from the Twitter images starts to determine the performance of both neural networks. The previously trained model only determines the start position of the fine-tuned model.

Meanwhile, for each model, we respectively select the top 5 positive and top 5 negative examples from the 1269 Twitter images according to the evaluation scores. \figurename~\ref{fig:neg} show those examples for each model. In both figures, each column contains the images for one model. A green solid box means the prediction label of the image agrees with the human label. Otherwise, we use a red dashed  box. The labels of top ranked images in both neural network models are all correctly predicted. However, the images are not all the same. This on the other hand suggests that even though the two models achieve similar results after fine-tuning, they may have arrived at somewhat different local optima due to the different starting positions, as well as the transfer learning process. For all the baseline models, it is difficult to say which kind of images are more likely to be correctly classified according to these images. However, we observe that there are several mistakenly classified images in common among the models using low-level features (the four rightmost columns in~\figurename~\ref{fig:neg}). Similarly, for Sentibank and Sentribute, several of the same images are also in the top ranked samples. This indicates that there are some common learned knowledge in the low-level feature models and mid-level feature models.

\section{Conclusions}
Visual sentiment analysis is a challenging and interesting problem. In this paper, we adopt the recent developed convolutional neural networks to solve this problem. We have designed a new architecture, as well as new training strategies to overcome the noisy nature of the large-scale training samples. Both progressive training and transfer learning inducted by a small number of confidently labeled images from the target domain have yielded notable improvements. The experimental results suggest that convolutional neural networks that are properly trained can outperform both classifiers that use predefined low-level features or mid-level visual attributes for the highly challenging problem of visual sentiment analysis. Meanwhile, the main advantage of using convolutional neural networks is that we can transfer the knowledge to other domains using a much simpler fine-tuning technique than those in the literature e.g., \cite{CVPR2010}.

It is important to reiterate the significance of this work over the state-of-the-art~\cite{siersdorfer2010analyzing,borth2013large,yuan2013sentribute}. We are able to directly leverage a much larger weakly labeled data set for training, as well as a larger manually labeled dataset for testing. The larger data sets, along with the proposed deep CNN and its training strategies, give rise to better generalizability of the trained model and higher confidence of such generalizability. In the future, we plan to develop robust multimodality models that employ both the textual and visual content for social media sentiment analysis. We also hope our sentiment analysis results can encourage further research on online user generated content.

We believe that sentiment analysis on large scale online user generated content is quite useful since it can provide more robust signals and information for many data analytics tasks, such as using social media for prediction and forecasting. In the future, we plan to develop robust multimodality models that employ both the textual and visual content for social media sentiment analysis. We also hope our sentiment analysis results can encourage further research on online user generated content.

\section{Acknowledgments}
This work was generously supported in part by Adobe Research. We would like to thank Digital Video and Multimedia (DVMM) Lab at Columbia University for providing the half million Flickr images and their machine-generated labels.
\bibliography{aaai-2015}

\begin{thebibliography}{}

\bibitem[\protect\citeauthoryear{Asur and Huberman}{2010}]{asur2010predicting}
Asur, S., and Huberman, B.~A.
\newblock 2010.
\newblock Predicting the future with social media.
\newblock In {\em WI-IAT}, volume~1,  492--499.
\newblock IEEE.

\bibitem[\protect\citeauthoryear{Bengio}{2012}]{bengio2012practical}
Bengio, Y.
\newblock 2012.
\newblock Practical recommendations for gradient-based training of deep
  architectures.
\newblock In {\em Neural Networks: Tricks of the Trade}. Springer.
\newblock  437--478.

\bibitem[\protect\citeauthoryear{Bollen, Mao, and
  Pepe}{2011}]{bollen2011modeling}
Bollen, J.; Mao, H.; and Pepe, A.
\newblock 2011.
\newblock Modeling public mood and emotion: Twitter sentiment and
  socio-economic phenomena.
\newblock In {\em ICWSM}.

\bibitem[\protect\citeauthoryear{Bollen, Mao, and
  Zeng}{2011}]{bollen2011twitter}
Bollen, J.; Mao, H.; and Zeng, X.
\newblock 2011.
\newblock Twitter mood predicts the stock market.
\newblock {\em Journal of Computational Science} 2(1):1--8.

\bibitem[\protect\citeauthoryear{Borth \bgroup et al\mbox.\egroup
  }{2013a}]{borth2013sentibank}
Borth, D.; Chen, T.; Ji, R.; and Chang, S.-F.
\newblock 2013a.
\newblock Sentibank: large-scale ontology and classifiers for detecting
  sentiment and emotions in visual content.
\newblock In {\em ACM MM},  459--460.
\newblock ACM.

\bibitem[\protect\citeauthoryear{Borth \bgroup et al\mbox.\egroup
  }{2013b}]{borth2013large}
Borth, D.; Ji, R.; Chen, T.; Breuel, T.; and Chang, S.-F.
\newblock 2013b.
\newblock Large-scale visual sentiment ontology and detectors using adjective
  noun pairs.
\newblock In {\em ACM MM},  223--232.
\newblock ACM.

\bibitem[\protect\citeauthoryear{\c{C}aglar G{\"u}l\c{c}ehre \bgroup et
  al\mbox.\egroup }{2013}]{GulcehreCPB13}
\c{C}aglar G{\"u}l\c{c}ehre; Cho, K.; Pascanu, R.; and Bengio, Y.
\newblock 2013.
\newblock Learned-norm pooling for deep neural networks.
\newblock {\em CoRR} abs/1311.1780.

\bibitem[\protect\citeauthoryear{Cire{\c{s}}an \bgroup et al\mbox.\egroup
  }{2011}]{cirecsan2011flexible}
Cire{\c{s}}an, D.~C.; Meier, U.; Masci, J.; Gambardella, L.~M.; and
  Schmidhuber, J.
\newblock 2011.
\newblock Flexible, high performance convolutional neural networks for image
  classification.
\newblock In {\em IJCAI},  1237--1242.
\newblock AAAI Press.

\bibitem[\protect\citeauthoryear{Davidov, Tsur, and
  Rappoport}{2010}]{davidov2010enhanced}
Davidov, D.; Tsur, O.; and Rappoport, A.
\newblock 2010.
\newblock Enhanced sentiment learning using twitter hashtags and smileys.
\newblock In {\em ICL},  241--249.
\newblock Association for Computational Linguistics.

\bibitem[\protect\citeauthoryear{Duan \bgroup et al\mbox.\egroup
  }{2012}]{CVPR2010}
Duan, L.; Xu, D.; Tsang, I.-H.; and Luo, J.
\newblock 2012.
\newblock Visual event recognition in videos by learning from web data.
\newblock {\em IEEE PAMI} 34(9):1667--1680.

\bibitem[\protect\citeauthoryear{Grangier, Bottou, and
  Collobert}{2009}]{grangier2009deep}
Grangier, D.; Bottou, L.; and Collobert, R.
\newblock 2009.
\newblock Deep convolutional networks for scene parsing.
\newblock In {\em ICML 2009 Deep Learning Workshop}, volume~3.
\newblock Citeseer.

\bibitem[\protect\citeauthoryear{Hamel and Eck}{2010}]{hamel2010learning}
Hamel, P., and Eck, D.
\newblock 2010.
\newblock Learning features from music audio with deep belief networks.
\newblock In {\em ISMIR},  339--344.

\bibitem[\protect\citeauthoryear{Hinton, Osindero, and
  Teh}{2006}]{hinton2006fast}
Hinton, G.~E.; Osindero, S.; and Teh, Y.-W.
\newblock 2006.
\newblock A fast learning algorithm for deep belief nets.
\newblock {\em Neural computation} 18(7):1527--1554.

\bibitem[\protect\citeauthoryear{Hinton}{2010}]{hinton2010practical}
Hinton, G.
\newblock 2010.
\newblock A practical guide to training restricted boltzmann machines.
\newblock {\em Momentum} 9(1):926.

\bibitem[\protect\citeauthoryear{Hu \bgroup et al\mbox.\egroup
  }{2013}]{hu2013unsupervised}
Hu, X.; Tang, J.; Gao, H.; and Liu, H.
\newblock 2013.
\newblock Unsupervised sentiment analysis with emotional signals.
\newblock In {\em WWW},  607--618.
\newblock International World Wide Web Conferences Steering Committee.

\bibitem[\protect\citeauthoryear{Jia}{2013}]{Jia13caffe}
Jia, Y.
\newblock 2013.
\newblock {Caffe}: An open source convolutional architecture for fast feature
  embedding.
\newblock \url{http://caffe.berkeleyvision.org/}.

\bibitem[\protect\citeauthoryear{Jin \bgroup et al\mbox.\egroup
  }{2010}]{jin2010wisdom}
Jin, X.; Gallagher, A.; Cao, L.; Luo, J.; and Han, J.
\newblock 2010.
\newblock The wisdom of social multimedia: using flickr for prediction and
  forecast.
\newblock In {\em ACM MM},  1235--1244.
\newblock ACM.

\bibitem[\protect\citeauthoryear{Joshi \bgroup et al\mbox.\egroup
  }{2011}]{joshi2011aesthetics}
Joshi, D.; Datta, R.; Fedorovskaya, E.; Luong, Q.-T.; Wang, J.~Z.; Li, J.; and
  Luo, J.
\newblock 2011.
\newblock Aesthetics and emotions in images.
\newblock {\em IEEE Signal Processing Magazine} 28(5):94--115.

\bibitem[\protect\citeauthoryear{Kavukcuoglu \bgroup et al\mbox.\egroup
  }{2010}]{kavukcuoglu2010learning}
Kavukcuoglu, K.; Sermanet, P.; Boureau, Y.-L.; Gregor, K.; Mathieu, M.; and
  LeCun, Y.
\newblock 2010.
\newblock Learning convolutional feature hierarchies for visual recognition.
\newblock In {\em NIPS}, ~5.

\bibitem[\protect\citeauthoryear{Krizhevsky, Sutskever, and
  Hinton}{2012}]{krizhevsky2012imagenet}
Krizhevsky, A.; Sutskever, I.; and Hinton, G.~E.
\newblock 2012.
\newblock Imagenet classification with deep convolutional neural networks.
\newblock In {\em NIPS}, ~4.

\bibitem[\protect\citeauthoryear{LeCun \bgroup et al\mbox.\egroup
  }{1989}]{lecun1989backpropagation}
LeCun, Y.; Boser, B.; Denker, J.~S.; Henderson, D.; Howard, R.~E.; Hubbard, W.;
  and Jackel, L.~D.
\newblock 1989.
\newblock Backpropagation applied to handwritten zip code recognition.
\newblock {\em Neural computation} 1(4):541--551.

\bibitem[\protect\citeauthoryear{LeCun \bgroup et al\mbox.\egroup
  }{1998}]{lecun1998gradient}
LeCun, Y.; Bottou, L.; Bengio, Y.; and Haffner, P.
\newblock 1998.
\newblock Gradient-based learning applied to document recognition.
\newblock {\em Proceedings of the IEEE} 86(11):2278--2324.

\bibitem[\protect\citeauthoryear{LeCun, Kavukcuoglu, and
  Farabet}{2010}]{lecun2010convolutional}
LeCun, Y.; Kavukcuoglu, K.; and Farabet, C.
\newblock 2010.
\newblock Convolutional networks and applications in vision.
\newblock In {\em ISCAS},  253--256.
\newblock IEEE.

\bibitem[\protect\citeauthoryear{Li \bgroup et al\mbox.\egroup
  }{2010}]{li2010micro}
Li, G.; Hoi, S.~C.; Chang, K.; and Jain, R.
\newblock 2010.
\newblock Micro-blogging sentiment detection by collaborative online learning.
\newblock In {\em ICDM},  893--898.
\newblock IEEE.

\bibitem[\protect\citeauthoryear{Liu \bgroup et al\mbox.\egroup
  }{2003}]{liu2003building}
Liu, B.; Dai, Y.; Li, X.; Lee, W.~S.; and Yu, P.~S.
\newblock 2003.
\newblock Building text classifiers using positive and unlabeled examples.
\newblock In {\em ICDM},  179--186.
\newblock IEEE.

\bibitem[\protect\citeauthoryear{Maas \bgroup et al\mbox.\egroup
  }{2011}]{maas2011learning}
Maas, A.~L.; Daly, R.~E.; Pham, P.~T.; Huang, D.; Ng, A.~Y.; and Potts, C.
\newblock 2011.
\newblock Learning word vectors for sentiment analysis.
\newblock In {\em ACL},  142--150.

\bibitem[\protect\citeauthoryear{Morency, Mihalcea, and
  Doshi}{2011}]{Morency:2011:TMS:2070481.2070509}
Morency, L.-P.; Mihalcea, R.; and Doshi, P.
\newblock 2011.
\newblock Towards multimodal sentiment analysis: Harvesting opinions from the
  web.
\newblock In {\em ICMI},  169--176.
\newblock New York, NY, USA: ACM.

\bibitem[\protect\citeauthoryear{O'Connor \bgroup et al\mbox.\egroup
  }{2010}]{o2010tweets}
O'Connor, B.; Balasubramanyan, R.; Routledge, B.~R.; and Smith, N.~A.
\newblock 2010.
\newblock From tweets to polls: Linking text sentiment to public opinion time
  series.
\newblock {\em ICWSM} 11:122--129.

\bibitem[\protect\citeauthoryear{Pang and Lee}{2008}]{pang2008opinion}
Pang, B., and Lee, L.
\newblock 2008.
\newblock Opinion mining and sentiment analysis.
\newblock {\em Foundations and trends in information retrieval} 2(1-2):1--135.

\bibitem[\protect\citeauthoryear{Patterson and
  Hays}{2012}]{Patterson2012SunAttributes}
Patterson, G., and Hays, J.
\newblock 2012.
\newblock Sun attribute database: Discovering, annotating, and recognizing
  scene attributes.
\newblock In {\em CVPR}.

\bibitem[\protect\citeauthoryear{Plutchik}{1984}]{plutchik1984emotions}
Plutchik, R.
\newblock 1984.
\newblock Emotions: A general psychoevolutionary theory.
\newblock {\em Approaches to emotion} 1984:197--219.

\bibitem[\protect\citeauthoryear{Siersdorfer \bgroup et al\mbox.\egroup
  }{2010}]{siersdorfer2010analyzing}
Siersdorfer, S.; Minack, E.; Deng, F.; and Hare, J.
\newblock 2010.
\newblock Analyzing and predicting sentiment of images on the social web.
\newblock In {\em ACM MM},  715--718.
\newblock ACM.

\bibitem[\protect\citeauthoryear{Tumasjan \bgroup et al\mbox.\egroup
  }{2010}]{tumasjan2010predicting}
Tumasjan, A.; Sprenger, T.~O.; Sandner, P.~G.; and Welpe, I.~M.
\newblock 2010.
\newblock Predicting elections with twitter: What 140 characters reveal about
  political sentiment.
\newblock {\em ICWSM}  178--185.

\bibitem[\protect\citeauthoryear{Yuan \bgroup et al\mbox.\egroup
  }{2013}]{yuan2013sentribute}
Yuan, J.; Mcdonough, S.; You, Q.; and Luo, J.
\newblock 2013.
\newblock Sentribute: image sentiment analysis from a mid-level perspective.
\newblock In {\em Proceedings of the Second International Workshop on Issues of
  Sentiment Discovery and Opinion Mining}, ~10.
\newblock ACM.

\bibitem[\protect\citeauthoryear{Zhang, Fuehres, and
  Gloor}{2011}]{zhang2011predicting}
Zhang, X.; Fuehres, H.; and Gloor, P.~A.
\newblock 2011.
\newblock Predicting stock market indicators through twitter ¡°i hope it is not
  as bad as i fear¡±.
\newblock {\em Procedia-Social and Behavioral Sciences} 26:55--62.

\end{thebibliography}
\bibliographystyle{aaai}
\end{document}